\documentclass[letterpaper, 10 pt, conference]{ieeeconf}  % Comment this line out
\IEEEoverridecommandlockouts                              % This command is only
\usepackage{amsmath}

\usepackage{subcaption}
\usepackage{makecell}
\usepackage{array}
\usepackage{graphics} % for pdf, bitmapped graphics files
\usepackage{amsmath} % assumes amsmath package installed
\usepackage{textcomp}
\usepackage{todonotes}
\usepackage{pdfpages}
\usepackage{url}
\usepackage{mathtools}
\usepackage{bm}
\usepackage{esvect}
\usepackage{float}
\usepackage{svg}
\usepackage{amsmath}
\usepackage{dblfloatfix}
\usepackage{siunitx}
\usepackage{graphicx,color,transparent}

\usepackage[utf8]{inputenc}
\usepackage[T1]{fontenc}
\usepackage{multirow}
 \DeclareMathOperator{\atantwo}{atan2}

\graphicspath{{.}}
\title{\LARGE \bf
LBGP: Learning Based Goal Planning for \\
Autonomous Following in Front
}

\author{Payam Nikdel$^1$, Richard Vaughan$^1$, Mo Chen$^1$
\thanks{$^1$School of Computing Science, Simon Fraser University, Canada. {\tt\small \{pnikdel, vaughan, mochen\}@sfu.ca}}}

\renewcommand{\baselinestretch}{.94} % Try not to go below .92

\begin{document}

\maketitle
\thispagestyle{empty}
\pagestyle{empty}

%%%%%%%%%%%%%%%%%%%%%%%%%%%%%%%%%%%%%%%%%%%%%%%%%%%%%%%%%%%%%%%%%%%%%%%%%%%%%%%%

\begin{abstract}

%  This paper investigates a learning based solution for the application of an autonomous robot which follows the user while staying ahead of them. Following in front is a challenging setting as the user's intended trajectory is unknown and needs to be estimated by the robot.   “Some intro sentece
% We combine x and y to ”

% our learning-based goal planner estimates human trajectory and produces  short navigational goala to guide the robot. These goals are used by a robotics trajectory planner to smoothly navigate the robot in front of user . We employ a curriculum learning approach to efficiently train the learning based module.
% %
% Our system outperforms the-state-of-the-art following in front and is more reliable compared to  End-to-End alternatives in both the simulation and real world experiments. Our results imply that this approach is directly transferable to real world and it can improve the generalizability and safety of an End-to-End approach.
%  --------------------

 This paper investigates a hybrid solution which combines deep reinforcement learning (RL) and classical trajectory planning for the ``following in front'' application.
 Here, an autonomous robot aims to stay ahead of a person as the person freely walks around.
 Following in front is a challenging problem as the user's intended trajectory is unknown and needs to be estimated, explicitly or implicitly, by the robot.
 In addition, the robot needs to find a feasible way to safely navigate ahead of human trajectory.
 %We combine deep Reinforcement Learning (RL) with a robotics trajectory planner to take advantage of both.
 Our deep RL module implicitly estimates human trajectory and produces short-term navigational goals to guide the robot.
 These goals are used by a trajectory planner to smoothly navigate the robot to the short-term goals, and eventually in front of the user.
 We employ curriculum learning in the deep RL module to efficiently achieve a high return.
 Our system outperforms the state-of-the-art in following ahead and is more reliable compared to  end-to-end alternatives in both the simulation and real world experiments.
 In contrast to a pure deep RL approach, we demonstrate zero-shot transfer of the trained policy from simulation to the real world.
%  Using a curriculum learning approach our system learn more efficiently while achieving a higher return.

% -----------

%  Our system outperforms the-state-of-the-art following in front or an End-to-End solution in both the simulation and real world.

%  ----------------

\end{abstract}

%%%%%%%%%%%%%%%%%%%%%%%%%%%%%%%%%%%%%%%%%%%%%%%%%%%%%%%%%%%%%%%%%%%%%%%%%%%%%%%%
\section{Introduction}

In many applications of human robot interaction a robot needs to stay near a user.
These include capturing the video of physical activities or monitoring elderly people.
Variants of following the user include  following from behind, following in front, and following side by side \cite{Ho2017, Hu2014DesignOS}.
Although following from behind is well studied, it can be much more challenging to stay ahead of a user \cite{Nikdel2018TheHP}.
To follow a user from behind, one can use a proportional integral derivative (PID) controller to keep the person at the center and add a separate PID controller to maintain person-robot distance \cite{leigh2015person}.
In contrast, for following in front, robot needs to explicitly or implicitly predict the future trajectory of the person and navigate to a point of that trajectory while maintaining a safe distance.
Behavioral experiments suggest that in following behind scenarios, the user frequently looks behind out of curiosity or to ensure the robot is within a safe distance \cite{jung2012control}.
Conversely, following in front can assist a person in different applications. Consider an autonomous shopping cart, self-driving luggage or autonomous guide dog; in all these applications, it is best if the robot is in front of the user.
The user not only feels safer, but also can interact with the robot more conveniently.

In recent years, a variety of research shows the capability of deep reinforcement learning (RL) to solve hard game problems \cite{silver2018general, berner2019dota}.
Applying it to robotics problems can help to address navigational tasks while considering the user intents \cite{chen2019relational, kulhanek2019vision}.
Deep RL can implicitly account for robot dynamics and enable a continuous interaction of the robot and its environment.
In the staying in front problem, deep RL can also implicitly predict a person's future trajectory, and continuously update the predictions to provide a smooth real-time experience for the user.
%As such, RL setup is an appealing choice for this study.

\begin{figure}[t]
	\centering
	\includegraphics[width=0.7\columnwidth]{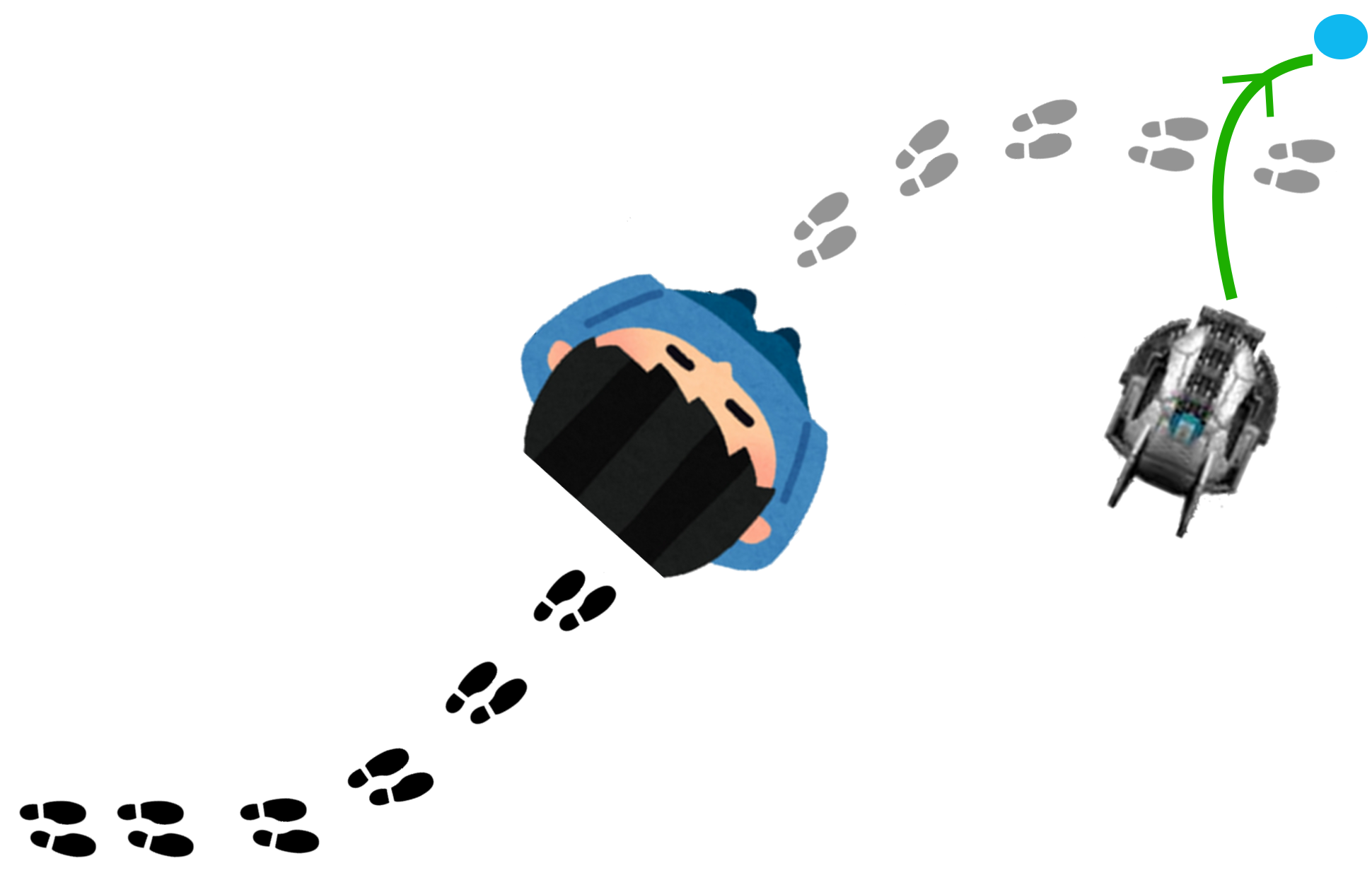}
    \caption{A mobile robot following-ahead of a user. The robot must predict the user's trajectory to stay in the correct position. In each step, robot considers previous states of the system to generate a goal (blue dot). Then a trajectory planner navigates the robot towards the goal (green line).}
    \label{fig:cool image}
	\vspace{-1.5em}
\end{figure}

In this paper, we propose the Learning Based Goal Planning (LBGP) approach to address the problem of staying in front of the user.
LBGP is a hybrid approach that uses the combination of deep RL and classical trajectory planners (see Figure \ref{fig:cool image}).
Our results show that combining deep RL with classical methods can greatly improve performance while maintaining its safety.
We also show the benefits of using curriculum learning to train the agent on increasingly challenging human motions.
Compared to training without a curriculum, our method trains the policy more efficiently while achieving a higher return.
%We start our curriculum with a straight user trajectory and gradually increase its complexity to walking in circle, smoothed curves and finally simulated trajectories of human.
To generalize our model to unseen and real-word inputs, we add a Gaussian noise to our observations.

We demonstrate favorable results in simulation and real world experiments compared to previous work.  Our ablation studies show the benefits of our hybrid approach and curriculum.
In particular, we show the effectiveness of our hybrid approach zero-shot transfer of the policy trained in simulation to the real world.
Example of our system can be find in the supplementary video \footnote{https://youtu.be/XSOUdPFPMmA}.
In summary, our main contributions are as follows:
\begin{itemize}
    \item We combine a robotics trajectory planner with deep RL to improve the safety and generalizability of our system.
    \item We use  curriculum learning to reduce the training time while improving the final return
    \item By evaluating our system in the simulation using a Clearpath Jackal and in the real world using a Turtlebot 2 robots, we show that our system can be more reliable and efficient for front-following in compare to an End-to-End or a hand-crafted approach
    \item We demonstrate that the policy trained using our method can directly transfer to the real world without any re-training.
\end{itemize}

\section{Related Work}
Person following has been studied for ground \cite{Nikdel2018TheHP, Pierre2018EndtoEndDL, Wang2018PersonDT}, aerial \cite{Huh2013IntegratedNS, Lugo2013FrameworkFA} or even underwater environments \cite{Islam2019TowardAG, Zadeh2018OnlinePP}.
Following from behind is the  dominant scenario in these studies.
Classical methods have divided the person following problem to a number of sub-modules: user localization, path finding and trajectory tracking \cite{leigh2015person, Wang2017Realtime3H}.
Learning navigational tasks directly from sensor inputs with End-to-End approaches has gained popularity in recent years \cite{Pierre2018EndtoEndDL}.
These techniques involve learning the task in simulation first and then possibly transfer the policy to the real word or generalize the policy to unseen environments \cite{Goldhoorn2014ContinuousRT}.
For a comprehensive review of autonomous person following, we encourage the reader to refer to \cite{Islam2019PersonfollowingBA}.

% -skydo folowing and capture a better perspective of the user (areial)
% Hu et al., 2014; Ferrer et al., 2013;
% The person can interact with the robot to make better decision.
\subsection{Following in Front}
% There are interesting scenarios where robot needs to stay ahead of person to accomplish the task.
Few papers had studied the interesting problem of following in front.
% Some methods depend a wearable tag by user in order to localize and stay ahead of them \cite{cifuentes2014human, tominaga2014around}. In this case, the robot can easily lose the sight of the person, especially in sharp turns.
Moustris et al. \cite{Moustris2016IntentionbasedFC} proposed a front-following system which incorporates a local dynamic planner along with a user intention recognition algorithm based on the user's relative offset from the middle of robot's view.
Ho et al. \cite{Ho2017} assumed a nonholomonic human model and used a Kalman filter to estimate human linear and angular velocity. They designed the robot motion planner such that it stays ahead of the user but the robot sometimes falls behind in specific situations like a T-junction. In a recent work, Nikdel et al. presented an Extended Kalman Filter (EKF) approach with joint 2D LiDAR and a fish-eye camera to detect and track the person \cite{Nikdel2018TheHP}. Their EKF model assumes a linear motion model. The EKF predicted position of the person  is corrected using a human motion model that consider obstacles.
To the best of our knowledge, this system is the state-of-the-art for the staying in front task for Unmanned Ground Vehicles (UGVs). We compare our LBGP with the result of this approach in an obstacle-free environment. Our approach shows a notable improvement.

\subsection{Reinforcement Learning}

%In recent years, RL received the attention of the robotics community.
%In robotics, it is costly or even infeasible to access annotated data.
To the best of our knowledge, the staying in front task has not been explored using an RL framework.
However, several studies used deep RL for related navigational tasks. Dewantara et al. proposed a guiding behaviour that optimize parameters of a social force model using Q-Learning \cite{socialaware}.
Recently, Chen et al. presented a relational graph deep RL approach for robotic crowd navigation \cite{chen2019relational}. Using these relational graph, they encoded higher-order interactions between agents and used it to anticipate future.  Curriculum learning approach has been used to increase the efficiency of RL training. Narvekar and Stone formulated a curriculum sequencing problem as a Markov Decision Process \cite{narvekar2019learning}. They show how curriculum learning can reduce training time.
Kulhanek et al. presented another RL based navigation agent \cite{kulhanek2019vision} which learns to navigate in an environment using only the raw images. They proposed to pre-train the network by transferring the learned policies from one environment to another and gradually increased the environment complexity.
We deploy a similar approach by  gradually increasing the difficulty of our person motion model.
Bansal et al. proposed a navigational framework for combining optimal control and learning \cite{bansal2020combining}. Their learning-based perception module produces a series of way-points that guides a robot toward the goal. One fundamental difference between our LBGT system and way point-based navigation approaches such as \cite{bansal2020combining,Li2019}  is the need to predict human trajectory which can make the task more challenging in comparison with navigating to a known goal.

\begin{figure}[h]
	\centering
	\includegraphics[width=0.6\columnwidth]{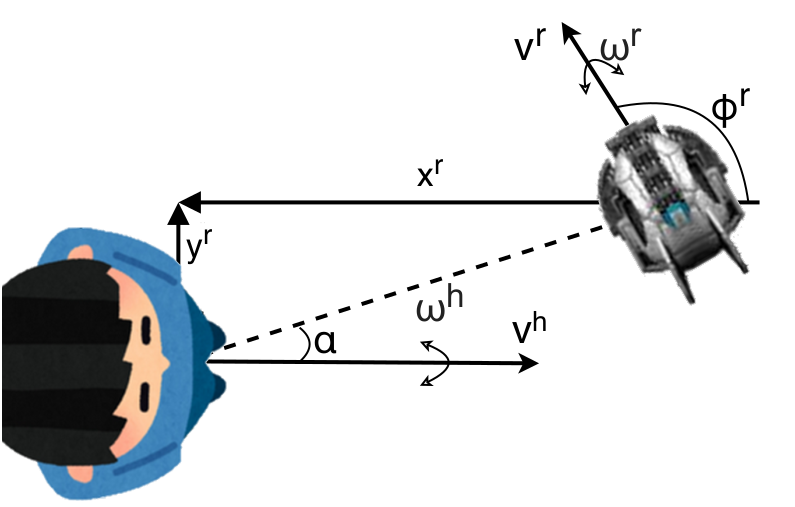}
	\caption{Our relative coordinates system}
    \label{fig:model_RL}
        \vspace{-5pt}

\end{figure}

\section{Problem Setup}
    In this work, we study the problem of keeping an autonomous robot in front of a walking person. We assume an obstacle free environment in which the robot should avoid collision with the human. We represent the global state of the human and robot with $(X_t^h, Y_t^h, \Phi_t^h, v_t^h, \omega_t^h )$, $(X_t^r, Y_t^r, \Phi_t^r, v_t^r, \omega_t^r)$, respectively. $(X_t, Y_t)$ is the position, $\Phi_t$ is the orientation, $v_t$ is the linear velocity and $\omega_t$ is the angular velocity at time t.

% For the dynamic of both the human and the robot we have:
% \begin{equation}
% \begin{pmatrix}
% \dot{X} \\
% \dot{Y} \\
% \dot{\Theta}
% \end{pmatrix} =
% \begin{pmatrix}
% cos(\Theta) & 0 & 0 \\
% 0 & sin(\Theta) & 0 \\
% 0 & 0 & 1
% \end{pmatrix} *
% \begin{pmatrix}
% v \\
% v \\
% \omega
% \end{pmatrix}
% \end{equation}

To make our approach transferable to real world and avoid over-fitting we use a relative state of the robot with respect to human, denoted $z_t^r$:

\begin{equation}
    z_t^r = (x_t^r, y_t^r, \varphi_t^r)
    \label{eq:observation_r}
\end{equation}

\[
\text{where }
\begin{bmatrix}
   x_t^r\\
    y_t^r \\
    \varphi^r_t
\end{bmatrix}
=
\begin{bmatrix}
    \cos(\Phi^h_t)       &-\sin(\Phi^h_t) & 0\\
    \sin(\Phi^h_t)       &\cos(\Phi^h_t)  & 0\\
    0 & 0 & 1
\end{bmatrix}
\begin{bmatrix}
    X_t^r-X_t^h \\
    Y_t^r -Y_t^h\\
    \Phi^r_t-\Phi^h_t
\end{bmatrix}
\]

For the purpose of calculating rewards, we define $\alpha_t = \atantwo{(y_t^r,x_t^r)}$ as the person-robot angle.

We also define a similar notation for the $i$th previous state of the human relative to their current state at time t:
\begin{equation}
    z_{t-i}^h: (x_{t-i}^h, y_{t-i}^h, \varphi_{t-i}^h)
    \label{eq:observationh}
\end{equation}

\[
\text{where}
\begin{bmatrix}
   x_{t-i}^h\\
    y_{t-i}^h \\
    \varphi^h_{t-i}
\end{bmatrix}
=
\begin{bmatrix}
    \cos(\Phi^h_t)       &-\sin(\Phi^h_t) & 0\\
    \sin(\Phi^h_t)       &\cos(\Phi^h_t)  & 0\\
    0 & 0 & 1
\end{bmatrix}
\begin{bmatrix}
    X_{t-i}^h-X_t^h \\
    Y_{t-i}^h -Y_t^h\\
    \Phi^h_{t-i}-\Phi^h_t
\end{bmatrix}
\]
As part of the observation, we consider a history of coordinates for both the robot and human. These coordinates are all respective to the latest position of the human. This relative system is visualized in Figure \ref{fig:model_RL}.

\section{Method}
 Our key insight in this paper is to combine an RL module with a classical trajectory planner. The agent uses our implementation of Deep Deterministic Policy Gradient (D4PG) \cite{d4pg} algorithm to generate a short-term navigational goal. A Time Elasic Bands (TEB) motion planner is used to navigate toward this goal, while treating the person as a dynamic obstacle. Our approach differs from typical policies trained with RL, which directly output an agent's actions.

\subsection{Observations and Navigational Goals}
The observation is a stack of robot and human relative states $(z^r_{t}, z^h_{t-1}, z^r_{t-1}, \ldots, z^h_{t-9}, z^r_{t-9})$, with $t$ being the current time step (see equations \eqref{eq:observation_r} and \eqref{eq:observationh}). We stack states up to the last 10 frames (at 5 FPS).
% where each contains linear velocity, angular velocity, relative position and relative orientation of robot or human.

 The numbers are continuous and will be scaled to $[-1, 1]$. In simulation, we capture all the variables from Gazebo simulator. In real world, it can be obtained by a motion capture system or human detection algorithms (e.g. YOLOv2 \cite{redmon2016yolo9000}) with RGB-D inputs (this approach was previously used in \cite{Nikdel2018TheHP}).
To improve the transferability of our approach to the real world, we add Gaussian noise to the observations in simulation.

The output of our policy network is a target position relative to the person. This position is a short-term navigational goal based on the estimated path of the user. We feed this output to the TEB local planner to navigate the robot in a smooth trajectory.

\begin{figure}[h]
	\centering
	\includegraphics[width=0.4\columnwidth]{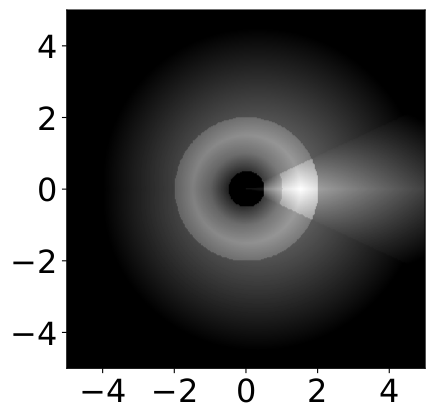}
	\caption{Reward based on the robot's relative position to the person. Increasing from black (-1) to white (+1). }
    \label{fig:reward}
        \vspace{-5pt}

\end{figure}
\label{subsection:reward}

\subsection{Reward}
We define the reward function such that the agent receives a higher reward (R) if it stays in front of the person at a desired distance of 1.5 m and negative reward if it is far away, too close or behind the person. The reward is scaled to $[-1, +1]$. Figure \ref{fig:reward} shows the reward function based on relative coordinates of the robot to human. The agent reward is defined as follows:

\[
    R_d=
\begin{cases}
    -1,& D < 0.5 \text{ or } 5 < D \\
    -(1 - D),& 0.5 < D < 1 \\
    0.5 (0.5 - |D - 1.5|),&  1.0 < D < 2\\
    -0.25 (D - 1)   &  2 < D < 5
    \end{cases}
\]
\[
    R_o=
\begin{cases}
    0.5 ((25 - |\alpha|) / 25) ,&|\alpha| < 25 \\
     -0.25 |\alpha| / 180, &  |\alpha| > 25 \\
\end{cases}
\]
\[
R = min(max(R_o + R_d, -1), 1)
\]

\noindent where $D$ is the distance between the robot and the person, and $\alpha$ is the angle between the person-robot vector and the person-heading vector (person-robot angle, in short).
We terminate the episode if the agent is too close ($D < 0.5$ m) or far away ($D > 5$ m).

\subsection{Policy Training Environment}
Our LBGP system is implemented in ROS \cite{quigley2009ros} and trained in the Gazebo robot simulator \cite{1389727}. We use a Turtlebot 3 burger robot as the person and a Clearpath Jackal robot as the robot. The person is controlled using our person motion model.
We design a world in the Gazebo simulator with four replicas of an environment each containing one learning agent. Three of agents explore the environment while the last one exploits the policy. This setup is arranged to mitigate the exploration exploitation trade-off. The simultaneously collected exposures are added to replay buffer to update the model weights.

\subsection{Curriculum Learning}
\label{subsection:person_motion}
\begin{figure}[h]
	\centering
	\includegraphics[width=0.7\columnwidth]{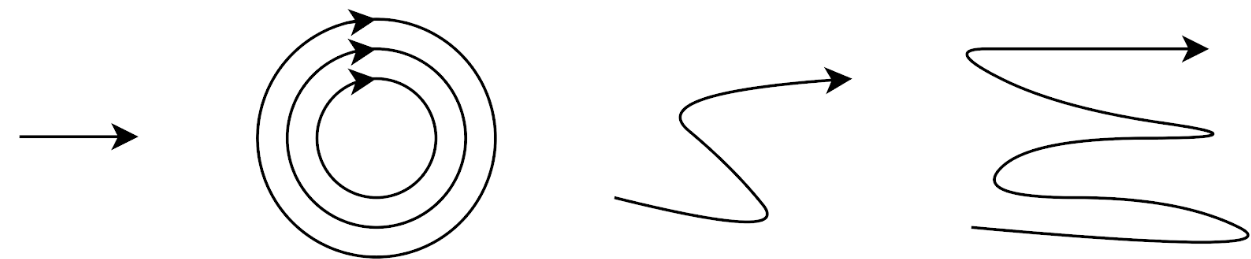}
	\caption{Visualization of person motion model. From left to write: moving straight, in different circles, in smoothed curves and using annotated simulated path of a human.}
    \label{fig:traj}
    \vspace{-5pt}
\end{figure}

To improve learning efficiency, we employ curriculum learning to train the agent in a series of tasks with increasing difficulty.
These tasks are defined based on the human trajectory.
We start with a straight line and move to more difficult trajectories as we go further in the training. In our curriculum, there are four difficulty levels: straight, circles, smoothed curves and simulated human trajectories explained below (see Figure \ref{fig:traj}). At each difficulty level, the robot is randomly spawned at positions between 1 to 2.5 meters away from the person with  uniformly random orientations.
The details of each difficulty level is elaborated below.

\subsubsection{\textbf{Straight}}
The person moves with an initial random linear velocity throughout the episode.

\subsubsection{\textbf{Circles}}
The person moves in a circle with a different radius each time. We create the circular motion by selecting a random initial linear velocity in $[0.2, 0.6]$ \SI{}{\meter/\second} and a random angular velocity in $[0.3, 0.8]$ \SI{}{rad/\second}.

\subsubsection{\textbf{Smoothed curves}}
The person moves in random curves generated by following linear velocities $V_l^t$ and angular velocities $V_a^t$:
 \vspace{-5pt}
 \[
 V_l^t = V_l^{t-1} - (V_l^{t-1} - R_1^{t-1})/3.
 \]
 \vspace{-5pt}
 \[
V_a^t = V_a^{t-1} - (V_a^{t-1} - R_2^{t-1})/3.
\]

where $R_1 , R_2$ are random numbers between $[0, 1]$ and $[-1, 1]$ respectively.

\subsubsection{\textbf{Simulated trajectories}}
We first arbitrarily ``draw'' trajectories by moving a robot using a joystick in Gazebo to cover the space while recording robot coordinate points. The total length of all  trajectories is roughly 50 meters.

During training, the person starts at a random point and tracks the above trajectories using a proportional integral derivative (PID) controller. To add variety to the data, we use 10 different environments and add the reverse of each trajectory to the library of trajectories as well.

\section{Simulation Experiment}
\label{section:simulation-experiment}
In this section, we present our experiments in simulation. We compare LBGP (our system) with two baselines: the latest Hand Crafted Following Ahead (HC) method in \cite{Nikdel2018TheHP} and an End-to-End learning Following Ahead (E2E) approach. HC system exploits EKF to predict the position of the user and then navigates to a point ahead of the predicted position using a trajectory planner. For sake of consistency, we use the same TEB motion planner as in HC.
For E2E approach, we use the same D4PG implementation with curriculum leaning, but instead of a navigational goal, the policy directly outputs the robot's linear and angular velocities scaled to $[-1, 1]$ m/s and $[-2, 2]$ rad/s, respectively.

% we scale the output to linear velocity of [-1, 1] m/s and angular velocity of [-2, 2] rad/s.

We conduct three experiments with different human trajectories.  In each experiment, we report the mean person-robot angle $\alpha$, the mean robot-user distance $D$ and the episode accumulated reward. The results of all the three experiments are included in Table \ref{tab:simulation_metrics}. In all experiments, the robot has no prior knowledge of the planned trajectory of the human.
%  The reward here is what we defined (see Section \ref{subsection:reward}) to measure the accuracy of following the person.
% The episode will be terminated in case of a collision or if the robot gets far away from the person (more than 5 meters).

\subsection{Straight}

\begin{table}[h]
  \centering
  \begin{tabular}{  c |l | r| r| r }

    Human & Approach & Distance  & Orientation & Reward \\
    Trajectory &  & mean $\pm$ std &  mean $\pm$ std &  \\\hline
    \multirow{4}{*}{  \begin{tabular}{@{}c@{}} Straight \\ ahead\end{tabular}}
    &&&&\\
    &LBGP & $1.53\pm0.2$ & $7.1\pm6.7$ & $28.31$ \\
    &HC & $1.35\pm0.2$ & $-1.1\pm1.6$ & $\bm{33.44}$ \\
    &E2E & $1.75\pm0.3$ & $8.9\pm7.2$ & $27.50$ \\

    \multirow{4}{*}{  \begin{tabular}{@{}c@{}} Straight \\ behind\end{tabular}}
    &&&&\\
    &LBGP & $1.59\pm0.2$ & $64.3\pm62.6$ & $\bm{10.50}$ \\
    &HC &  $1.10\pm0.2$ & $86.1\pm66.6$ & $1.30$ \\
    &E2E & $1.54\pm0.5$ & $91.8\pm63.1$ & $-3.13$ \\

    \multirow{4}{*}{  \begin{tabular}{@{}c@{}} Turning \\ ahead\end{tabular}}
    &&&&\\
    &LBGP & $1.90\pm0.3$ & $-5.7\pm10.3$ & \bm{$24.96$} \\
    &HC &  $1.04\pm0.2$ & $-12.6\pm11.6$ & $20.20$ \\
    &E2E & $1.96\pm0.3$ & $-7.5\pm4.6$ & $22.44$ \\

    \multirow{4}{*}{  \begin{tabular}{@{}c@{}} Turning \\ behind\end{tabular}}
    &&&&\\
    &LBGP & $1.69\pm0.3$ & $81.7\pm67.8$ & \bm{$0.40$} \\
    &HC &  $1.07\pm0.3$ & $83.9\pm83.4$ & $-5.03$ \\
    &E2E &  $1.62\pm0.5$ & $55.1\pm81.1$ & $0.17$ \\

    \multirow{4}{*}{  \begin{tabular}{@{}c@{}} Turning \\ inside\end{tabular}}
    &&&&\\
    &LBGP &  $1.72\pm0.3$ & $1.6\pm23.6$ & \bm{$23.04$} \\
    &HC &  $0.98\pm0.3$ & $3.6\pm24.1$ & $8.37$ \\
    &E2E & $2.11\pm0.5$ & $-1.5\pm18.7$ & $11.36$ \\

    \multirow{4}{*}{  \begin{tabular}{@{}c@{}} Turning \\ outside\end{tabular}}
    &&&&\\
    &LBGP & $1.56\pm0.2$ & $-33.1\pm19.8$ & \bm{$13.97$} \\
    &HC &  $1.07\pm0.2$ & $-22.8\pm18.7$ & $13.68$\\
    &E2E &  $1.32\pm0.2$ & $56.3\pm72.1$ & $2.53$ \\

    \multirow{4}{*}{  \begin{tabular}{@{}c@{}} Trajectory \\ one\end{tabular}}
    &&&&\\
    &LBGP & $1.37\pm0.3$ & $10.4\pm15.4$ & $\bm{27.17}$ \\
    &HC & $1.09\pm0.3$ & $-8.8\pm34.8$ & $0.91$ \\
    &E2E & $1.65\pm0.2$ & $22.9\pm42.4$ & $16.60$ \\
    \multirow{4}{*}{  \begin{tabular}{@{}c@{}} Trajectory \\ two\end{tabular}}
    &&&&\\
    &LBGP & $1.54\pm0.2$ & $-4.7\pm62.2$ & $\bm{15.86}$ \\
    &HC &  $1.12\pm0.4$ & $-57.0\pm45.8$ & $-5.83$ \\
    &E2E & $1.62\pm0.2$ & $-1.7\pm72.4$ & $10.99$ \\

    \multirow{4}{*}{  \begin{tabular}{@{}c@{}} Trajectory \\ three\end{tabular}}
    &&&&\\
    &LBGP & $1.59\pm0.3$ & $12.6\pm81.0$ & $\bm{11.79}$ \\
    &HC &  $1.15\pm0.4$ & $6.2\pm75.8$ & $-8.02$ \\
    &E2E & $1.92\pm0.3$ & $11.0\pm80.1$ & $2.15$ \\

  \end{tabular}

  \caption{Comparison of our systems versus two baselines for all simulation trajectories. }\label{tab:simulation_metrics}
      \vspace{-5pt}

\end{table}

The first experiment conducted on straight human motion trajectory to compare the behaviour of the three methods. The human simply start moving forward with a constant linear velocity of 0.6 m/s. We spawn the robot relative to person in two initial settings, \textbf{Ahead}: ($D=1.5$m, $\alpha=0^\circ$) and \textbf{Behind}: ($D=1.5 $m, $\alpha=180^\circ$).

We compare our results with HC and E2E (Table \ref{tab:simulation_metrics}). For the Ahead setting, HC achieves the highest episode reward. HC can achieve a better results as the incorporated EKF relies on linearity of human motion and it can optimally follow the straight line. In LBGP training, we apply Gaussian noise thus the robot may slightly deviate to the sides. In the Behind setting, our approach achieves the highest reward as it has learned to keep a safe distance with the human by setting further navigational goals for TEB compared to HC.
% and exploits the usage of a trajectory planner to get in front as fast as possible .

\subsection{Turning}
We assess different approaches with turning trajectories. In this case, the person moves with a linear velocity of   0.3 m/s and angular velocity of 0.3 rad/s. To cover a large variety of initial conditions, we evaluate four positions of the robot relative to person, \textbf{Ahead}: ($D=1.5  $m, $\alpha=180^\circ$), \textbf{Behind}: ($D=1.5  $m, $\alpha=0^\circ$),   \textbf{Ahead-inside-the-turn}: ($D=1.5  $m, $\alpha=45^\circ$) or  \textbf{Ahead-outside-the-turn}: ($D=1.5  $m, $\alpha=-45^\circ$).  The result of this experiment shows that our LBGP achieves the highest reward in all the settings (Table \ref{tab:simulation_metrics}).

\subsection{Simulated trajectories}
We designed three simulated trajectories to further evaluate our system. Similar to the training phase, we employ PID controllers for the simulated human to follow totally unseen trajectories, and the learning agent attempts to stay in front of the simulated human. Figure \ref{fig:simulate_traj_sim} shows the robot's trajectories corresponding to the three different human trajectories. In this experiment, we can see a more noticeable difference in performance of LBGP compared to both baselines (Table \ref{tab:simulation_metrics}). Compared to E2E, our method likely performs better as the usage of a trajectory planner abstracts navigation to predicting a goal, while the E2E method needs to implicitly learn the dynamics of the system.

Learning accurate dynamics can be challenging and may expose the E2E to over-fitting.  Our method, LBGP, also outperforms the HC system, since LBGP predicts a goal based on a history of the human trajectory as opposed to using a linear human motion model as in HC.

abstract lower level control by directly predicting
\begin{table}[h]
  \centering
  \begin{tabular}{c|c |c| c }
      &Trajectory one &  Trajectory two & Trajectory three \\
                \Xhline{2\arrayrulewidth}
      LBGP&
    \begin{minipage}{.17\linewidth}
        \vspace{1pt}
      \includegraphics[width=\linewidth]{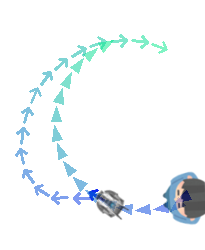}
    \end{minipage}
    &\begin{minipage}{.17\linewidth}
      \includegraphics[width=\linewidth]{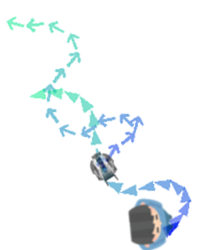}
    \end{minipage}
    &\begin{minipage}{.17\linewidth}
      \includegraphics[width=\linewidth]{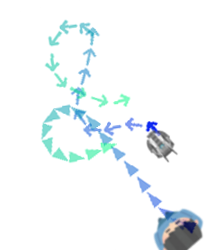}
    \end{minipage}\\
    E2E&
    \begin{minipage}{.17\linewidth}
      \includegraphics[width=\linewidth]{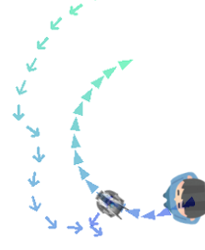}
    \end{minipage}
    &\begin{minipage}{.17\linewidth}
      \includegraphics[width=\linewidth]{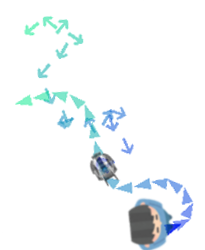}
    \end{minipage}
    &\begin{minipage}{.17\linewidth}
      \includegraphics[width=\linewidth]{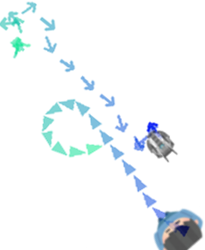}
    \end{minipage}\\
    HC&
    \begin{minipage}{.17\linewidth}
      \includegraphics[width=\linewidth]{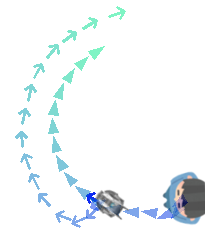}
    \end{minipage}
    &\begin{minipage}{.17\linewidth}
      \includegraphics[width=\linewidth]{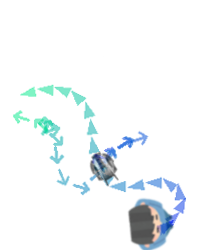}
    \end{minipage}
    &\begin{minipage}{.17\linewidth}
      \includegraphics[width=\linewidth]{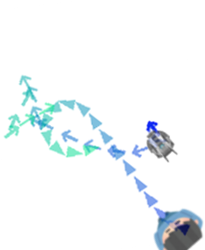}
    \end{minipage}\\

  \end{tabular}
  \captionof{figure}{Visualize the trajectory of robot (arrows) and human (triangle) during the simulated trajectory experiment for our system (LBGP) and two baselines, HC and E2E.}\label{fig:simulate_traj_sim}
      \vspace{-5pt}

\end{table}

\subsection{Ablation Study}
\begin{figure}[h]
	\centering
	\includegraphics[width=0.85\columnwidth]{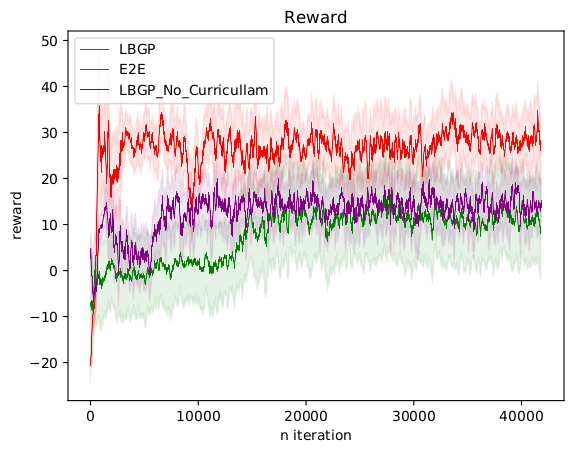}
	\caption{The shaded area represents half a standard deviation.}
    \label{fig:rewardcomp}
        \vspace{-5pt}

\end{figure}

We perform an ablation study to evaluate the effectiveness of different modules and training procedures of our LBGP approach.
We compare the performance of our approach to two variants of our it: 1) without curriculum learning (\emph{LBGP-no-curriculum}), and 2) without a trajectory planner (\emph{E2E}, same as described in Section \ref{section:simulation-experiment}). As shown in Figure \ref{fig:rewardcomp}, both \emph{planner-no-curriculum} and \emph{E2E} have slow learning curves and reach a lower discounted cumulative reward bound compared to LBGP, our proposed method.

\section{real world Experiment}
We test LBGP on a TurtleBot 2 hardware testbed, and evaluate the transferability of the policy trained in simulation, using our approach, to the real world. We also compare our method's sim2real ability with the two baselines, HC and E2E, defined on Section \ref{section:simulation-experiment}. We performed three experiments each with 4 different initial relative states. In short, our approach demonstrates successful zero-shot sim2real transfer of the policy.

To keep the experiments consistent between different approaches, all the initial states of human and robot along with the trajectory of human are marked on the ground with color tapes. In each experiment, we report the total discounted cumulative reward, the mean person-robot angle  ($\alpha$) and the mean person-robot distance ($D$) as a measures of the follow-ahead quality. To make the accumulated reward a fair evaluation, we keep a constant number of time step for each setting. We use a motion capture system to record the robot and person's states. For all the experiments we use the policy we trained in simulation with no changes. In each setting, we terminated the experiment as soon as the robot hits the person or gets more than three meters away from the person. As with the simulations, in every real world experiment, the robot has no knowledge of the planned trajectory of the human.

\subsection{\emph{Straight} Trajectory}
In this experiment, the initial positions of robot relative to human are:  \textbf{Ahead}: ($D=2 $m, $\alpha=0^\circ,$),  \textbf{Ahead-right}: ($D=1.5 $m, $\alpha=45^\circ$ ),  \textbf{Ahead-left}: ($D=1.5$m, $\alpha=-45^\circ$) and \textbf{Behind}: ($D=1.2$m, $\alpha=180^\circ$).  In each setting, the person intends to navigate with a constant forward speed toward a goal located at 7 meters of its initial position. The four settings along with the result of the \emph{Straight} experiment is reported in Table \ref{tab:straight_real}. In this experiment the EKF model of HC can correctly predict the human trajectory and it achieves the highest reward only for the Ahead setting. Among all the other settings, our LBGP method achieves the highest performance. It is likely because the policy in LBGP is trained to keep the safety distance with the human. E2E failed to accomplish the following ahead task due to collision with person (Ahead and Ahead-right settings) or drifting away in reverse direction (Behind setting). Likely, E2E learns to navigate only with the specific simulated robot dynamics and unable to generalize to a new robot dynamics in the real world experiments.
% change of the hardware (different robot in simulation and real-world) and a new robot dynamics in real world.

% It collided with the person in first and third settings and go to a reverse direction in the last setting.

\begin{table}[h]
  \centering
  \begin{tabular}{ c |l | r| r| r }
    % \hline
    Human & Approach & Distance  & Orientation & Reward \\
     Trajectory&  & mean $\pm$ std &  mean $\pm$ std  &  \\\hline
    \multirow{5}{*}{\begin{minipage}{.14\linewidth}
      \includegraphics[width=0.95\linewidth]{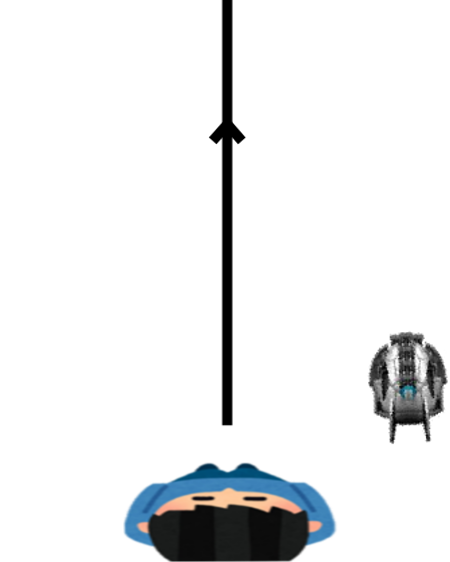}
    \end{minipage}}
    & & & &\\
    &LBGP & $1.63\pm0.4$ & $-8.4\pm26.6$ & \bm{$27.42$} \\
    &HC &  $1.24\pm0.6$ & $-11.8\pm16.4$ & $26.31$ \\
    &E2E & Failed & Failed & Failed \\
    &&&&\\
    \hline

    \multirow{5}{*}{\begin{minipage}{.14\linewidth}
      \includegraphics[width=0.95\linewidth]{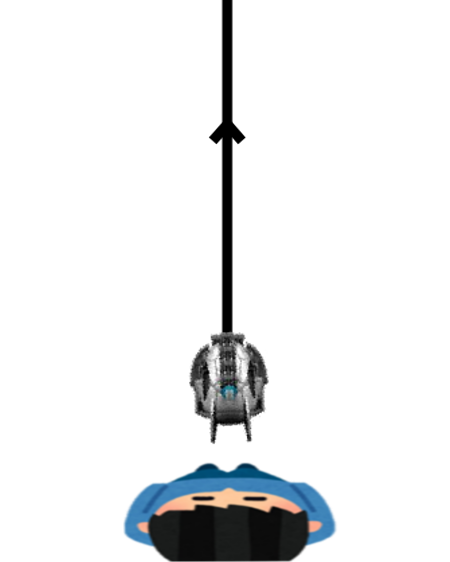}
    \end{minipage}}
    & & & &\\
    &LBGP & $1.24\pm0.3$ & $2.1\pm14.5$ & $40.92$ \\
    &HC & $1.14\pm0.2$ & $0.5\pm3.0$ & \bm{$61.08$} \\
    &E2E & $1.14\pm0.3$ & $-0.1\pm10.0$ & $41.59$\\
     &&&&\\
    \hline
    \multirow{5}{*}{\begin{minipage}{.14\linewidth}
      \includegraphics[width=0.95\linewidth]{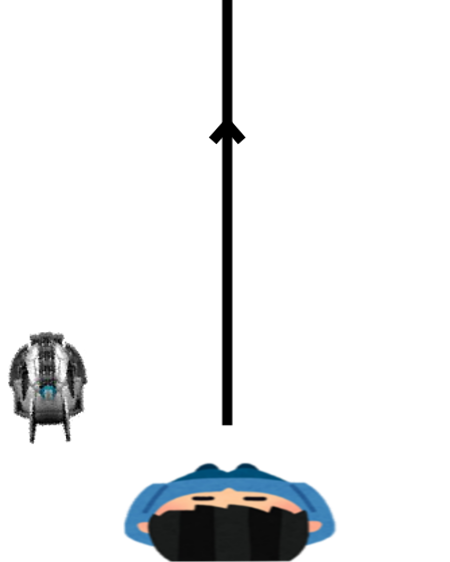}
    \end{minipage}}
    & & & &\\
    &LBGP & $1.61\pm0.4$ & $30.2\pm34.6$ & \bm{$15.68$} \\
    &HC & $1.28\pm0.7$ & $16.8\pm21.6$ & $14.24$ \\
    &E2E  & Failed & Failed & Failed \\ &&&&\\
    \hline
    \multirow{5}{*}{\begin{minipage}{.14\linewidth}
      \includegraphics[width=0.95\linewidth]{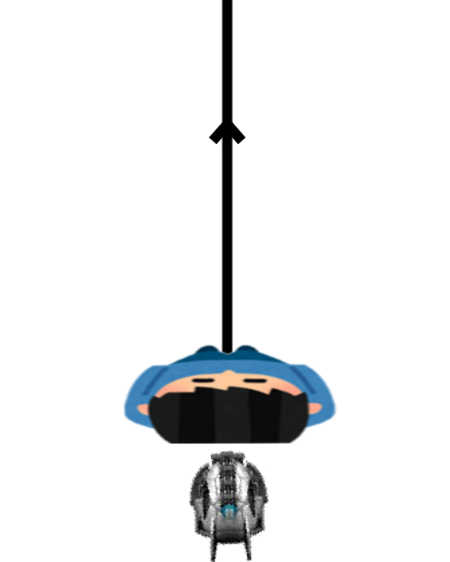}
    \end{minipage}}
    & & & &\\
    &LBGP & $1.85\pm0.3$ & $34.0\pm95.2$ & \bm{$-6.79$} \\
    &HC & $1.73\pm0.3$ & $-105.7\pm51.0$ & $-15.69$ \\
    &E2E  & Failed & Failed & Failed \\ &&&&\\
    % \hline

  \end{tabular}

  \caption{Comparison of our systems versus two baselines for straight trajectory.}\label{tab:straight_real}
      \vspace{-5pt}

\end{table}

\subsection{\emph{S shaped} Trajectory}
In the second experiment, we evaluate our system for an \emph{S} shaped trajectory. The initial relative position of robot is exactly similar to the \emph{Straight} trajectory. The user deliberately follows a \emph{S shape} path for all the settings.  As shown in Table \ref{tab:s_real}, LBGP achieves the highest reward in all four settings. When the person travels along an \emph{S shaped} trajectory, it is important to consider a history of the person to predict its future trajectory and a simple EKF as in HC cannot correctly predict the complexity of this motion. Figure \ref{fig:real_traj} visualizes examples of the robot and human trajectories for Ahead-right and Behind settings. Similar to the \emph{Straight} experiment, E2E failed three out of four settings by colliding with the user.

\begin{table}[h]
  \centering
  \begin{tabular}{ c |l | r| r| r }
    Human & Approach & Distance & Orientation &  Reward \\
     Trajectory&  & mean $\pm$ std &  mean $\pm$ std &  \\\hline
    \multirow{5}{*}{\begin{minipage}{.14\linewidth}
      \includegraphics[width=0.95\linewidth]{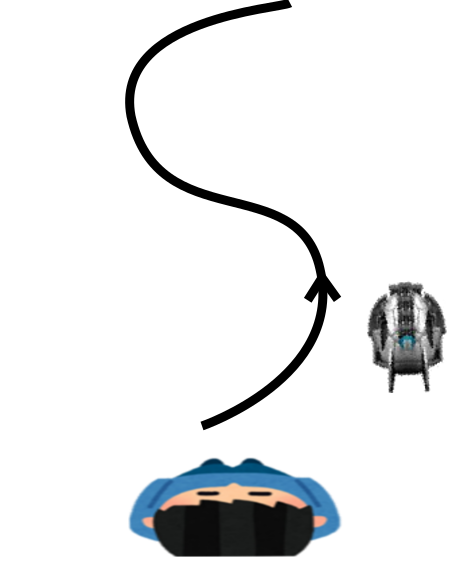}
    \end{minipage}}
    & & & &\\
    &LBGP &  $2.09\pm0.3$ & $-4.9\pm26.7$ & \bm{$2.30$}\\
    &HC &    $1.14\pm0.5$ & $-14.2\pm78.4$ & $-9.42$\\
    &E2E  & Failed & Failed & Failed \\
    &&&&\\
    \hline

    \multirow{5}{*}{\begin{minipage}{.14\linewidth}
      \includegraphics[width=0.95\linewidth]{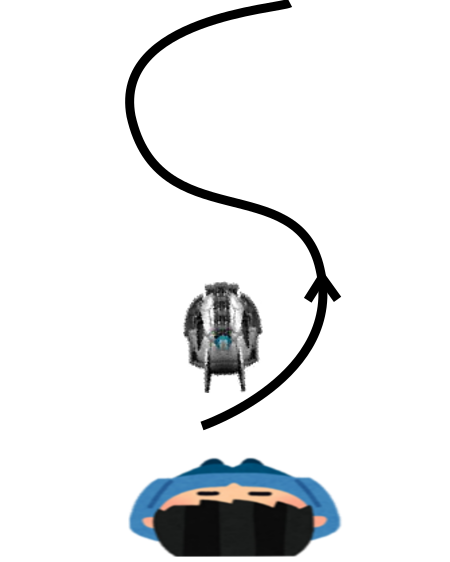}
    \end{minipage}}
    & & & &\\
    &LBGP & $1.86\pm0.4$ & $16.9\pm28.3$ & \bm{$5.15$}\\
    &HC & $1.04\pm0.3$ & $38.9\pm60.3$ & $-6.83$\\
    &E2E & $1.41\pm0.4$ & $-35.4\pm53.0$ & $3.45$\\
    &&&&\\
    \hline
    \multirow{5}{*}{\begin{minipage}{.14\linewidth}
      \includegraphics[width=0.95\linewidth]{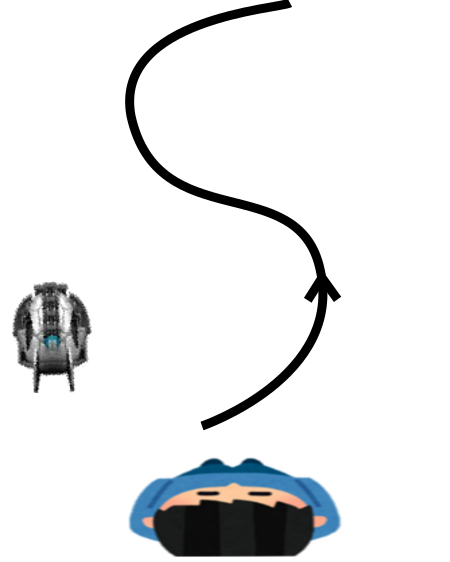}
    \end{minipage}}
    & & & &\\
    &LBGP & $1.81\pm0.6$ & $35.8\pm33.1$ & \bm{$11.74$} \\
    &HC & $1.60\pm0.8$ & $65.3\pm50.0$ & $-9.29$\\
    &E2E & Failed & Failed & Failed \\ &&&&\\
    \hline
    \multirow{5}{*}{\begin{minipage}{.14\linewidth}
      \includegraphics[width=0.95\linewidth]{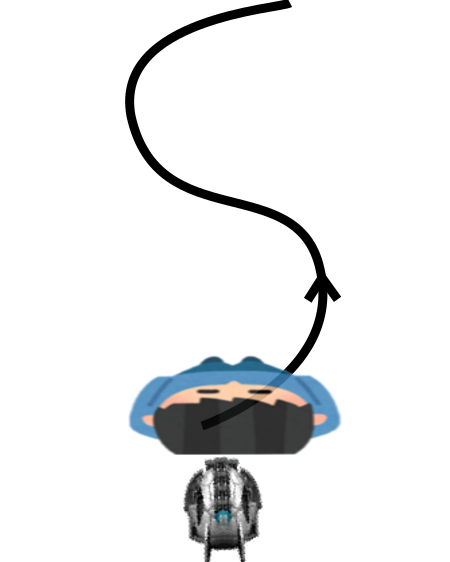}
    \end{minipage}}
    & & & &\\
    &LBGP &$1.30\pm0.2$ & $38.2\pm70.9$ & \bm{$7.84$}\\
    &HC & $1.69\pm0.4$ & $12.2\pm151.6$ & $-18.64$ \\
    &E2E  & Failed & Failed & Failed \\ &&&&\\

  \end{tabular}

  \caption{Comparison of our systems versus two baselines for "S shape" trajectory.}\label{tab:s_real}
      \vspace{-5pt}

\end{table}

\subsection{\emph{U-turn} Trajectory}
Lastly, we evaluate the LBGP when the person perform a U-turn. The initial positions of robot relative to human is: \textbf{Ahead}: ($D=1.7$m,  $\alpha=0^\circ$), \textbf{Ahead-left} ($D=2 $m, $\alpha=-55^\circ$), \textbf{Ahead-far-left} ($D=3.6$m, $\alpha=-75^\circ$) and \textbf{Behind} ($D=1.2$m, $\alpha=180^\circ$). Table \ref{tab:u_reall} shows the four settings along with the result of the \emph{U-turn} experiment, and LBGP consistently accumulates the highest reward. For a challenging U-turn trajectory, it is important for the robot to ``notice'' these specific walking patterns and react  spontaneously. This cannot be done in HC method as HC anticipate future based on the heading of the person. Examples of robot and human trajectories for  Ahead-left and Behind settings is visualized in Figure \ref{fig:real_traj}. For instance, in the Ahead-left setting, LBGP predict the turn early and avoid getting far away from the person. On the other hand, E2E has trouble transferring the policy to the real world.

\begin{table}[h]
  \centering
  \begin{tabular}{ c |l | r| r| r }

    Human & Approach & Distance & Orientation & Reward \\
     Trajectory&  & mean $\pm$ std &  mean $\pm$ std &  \\\hline
    \multirow{5}{*}{\begin{minipage}{.14\linewidth}
      \includegraphics[width=0.95\linewidth]{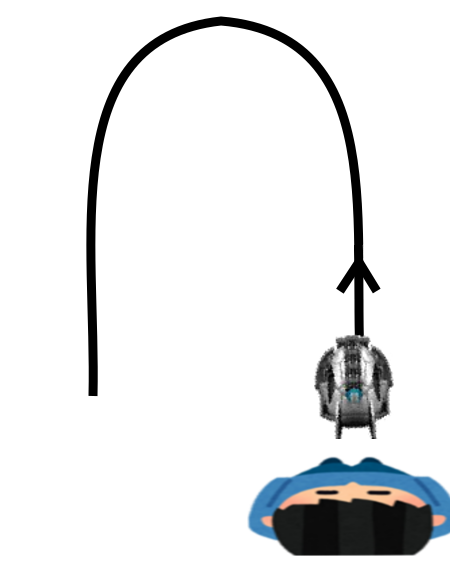}
    \end{minipage}}
    & & & &\\
    &LBGP & $1.99\pm0.2$ & $-16.8\pm18.6$ & \bm{$14.91$}\\
    &HC & $1.01\pm0.5$ & $-55.2\pm57.0$ & $-11.33$\\
    &E2E  & Failed & Failed & Failed \\
    &&&&\\
    \hline

    \multirow{5}{*}{\begin{minipage}{.14\linewidth}
      \includegraphics[width=0.95\linewidth]{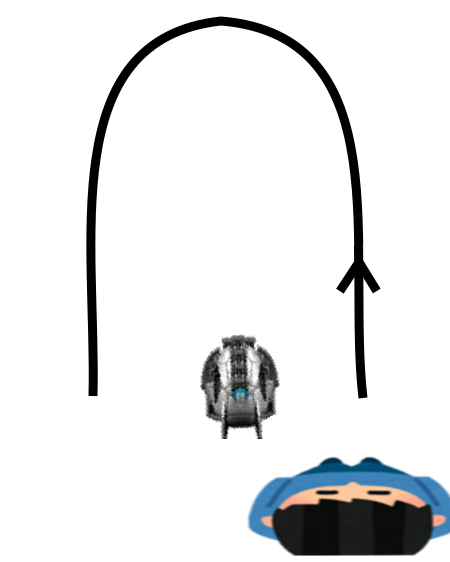}
    \end{minipage}}
    & & & &\\
    &LBGP & $1.34\pm0.4$ & $20.9\pm37.8$ & \bm{$18.73$}\\
    &HC & $1.15\pm0.3$ & $-28.7\pm98.1$ & $-8.09$ \\
    &E2E & $1.75\pm0.4$ & $43.3\pm26.7$ & $10.63$  \\
    &&&&\\
    \hline
    \multirow{5}{*}{\begin{minipage}{.14\linewidth}
      \includegraphics[width=0.95\linewidth]{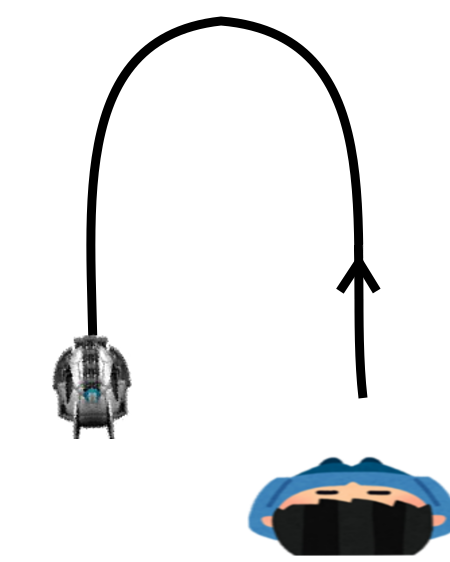}
    \end{minipage}}
    & & & &\\
    &LBGP & $1.91\pm0.7$ & $36.2\pm36.3$ & \bm{$16.52$}\\
    &HC &  $1.31\pm0.9$ & $-6.1\pm43.8$ & $-13.14$\\
    &E2E  & Failed & Failed & Failed \\
    &&&&\\
    \hline
    \multirow{5}{*}{\begin{minipage}{.14\linewidth}
      \includegraphics[width=0.95\linewidth]{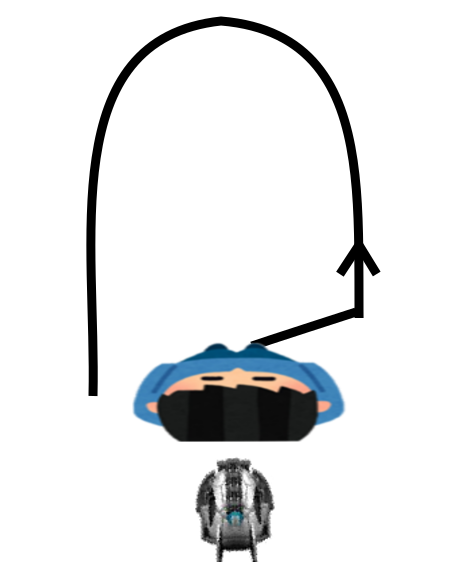}
    \end{minipage}}
    & & & &\\
    &LBGP & $1.53\pm0.3$ & $44.5\pm58.5$ & \bm{$20.03$}\\
    &HC & $1.12\pm0.3$ & $-1.7\pm120.3$ & $-11.83$ \\
    &E2E &$1.82\pm0.3$ & $67.7\pm45.4$ & $-7.43$\\
    &&&&\\

  \end{tabular}

  \caption{Comparison of our systems versus two baselines for U-turn trajectory.}\label{tab:u_reall}
      \vspace{-5pt}

\end{table}

\begin{table}[h]
  \centering
  \begin{tabular}{c|c }
      LBGP& HC\\
                \Xhline{2\arrayrulewidth}

    \begin{minipage}{.32\linewidth}
        \vspace{1pt}
      \includegraphics[width=\linewidth]{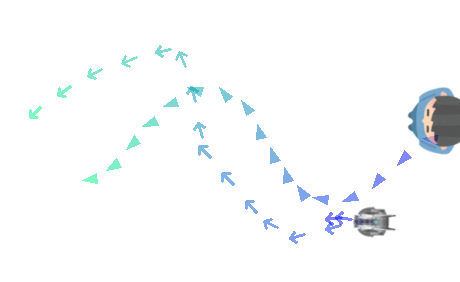}
    \end{minipage}
    &\begin{minipage}{.32\linewidth}
      \includegraphics[width=\linewidth]{images/smoothed_1_base.png}
    \end{minipage}\\
        \begin{minipage}{.32\linewidth}
        \vspace{1pt}
      \includegraphics[width=\linewidth]{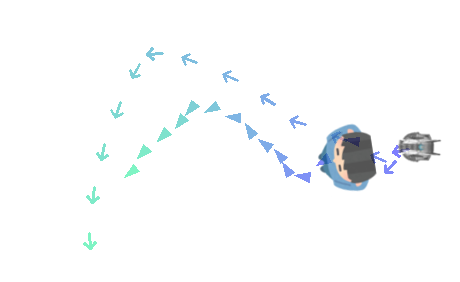}
    \end{minipage}
    &\begin{minipage}{.32\linewidth}
      \includegraphics[width=\linewidth]{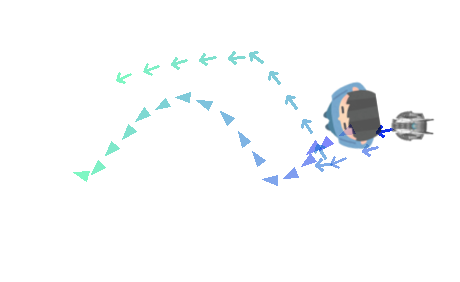}
    \end{minipage}\\

    \begin{minipage}{.32\linewidth}
        \vspace{1pt}
      \includegraphics[width=\linewidth]{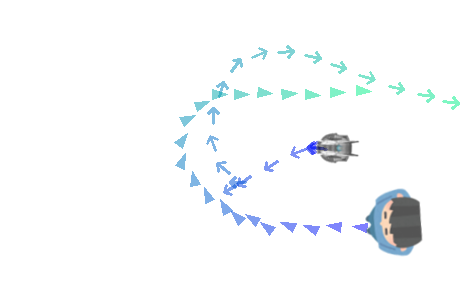}
    \end{minipage}
    &\begin{minipage}{.32\linewidth}
      \includegraphics[width=\linewidth]{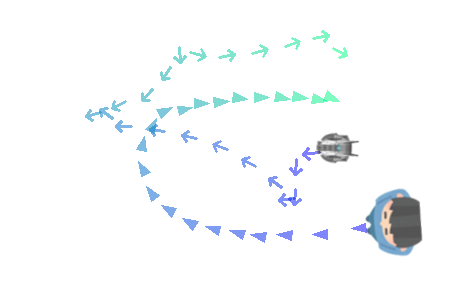}
    \end{minipage}\\

    \begin{minipage}{.32\linewidth}
        \vspace{1pt}
      \includegraphics[width=\linewidth]{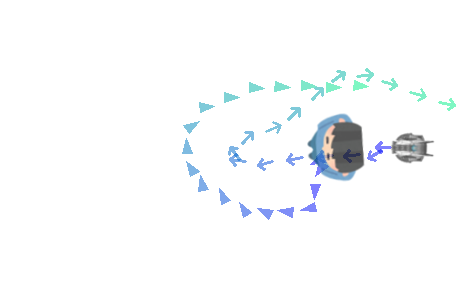}
    \end{minipage}
    &\begin{minipage}{.32\linewidth}
      \includegraphics[width=\linewidth]{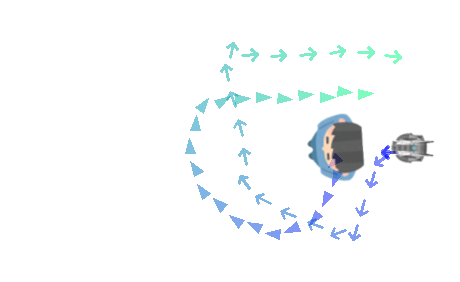}
    \end{minipage}\\

  \end{tabular}
  \captionof{figure}{real world Examples: the robot (in arrows) and user (in triangles) trajectories is depicted. Row 1 and 2, \emph{S shape} experiment in \emph{ahead-right} and \emph{behind} settings. Row 3 and 4: \emph{U-turn} experiment in \emph{ahead-left} and \emph{behind} settings.    }\label{fig:real_traj}
      \vspace{-5pt}

\end{table}

\section {Discussion}
\subsection{Comparison to the Hand Crafted method}
Our results show that our proposed learning-based system for following ahead outperform the HC method in both the simulation and real world. LBGP is able to create a complex model of environment with a better abstraction of human motion model as opposed to a linear EKF in the HC. Another advantage of RL is the large amount of training data that can be obtained in a simulated environment. This allows LBGP to better predict human trajectories (implicitly) compared to a hand-crafted method.

% Unless one consider each one of the trajectories and design a system for each one.
\subsection{Comparison to the End-to-End method}
Although E2E achieves a comparable performance in simulation, it is unreliable in the real world. Using E2E, robot collided with the user in all the three real world experiments.
We also saw robot shaking a lot when we use E2E. After investigating we identified the dynamics of the real world robot differs from the simulated one, which prevents E2E from extending the learned behaviour to the real world. In contrast, LBGP overcomes this model mismatch by abstracting away the dynamics using the TEB trajectory planner. This planner also helps our system to avoid any collision with person while staying in a safe distance.

% it is probably due to accuracy of robot model in the simulation. It can also be because of hardware difference. In simulation for all the training we use Jackal robot and in real world we use Turtlebot 2.
% \par
% Using a trajectory planner helps our LBGP to navigate in real world using different hardware safely. By using a combination of RL and a classical robotic trajectory planner we learn the human behaviour while keeping the safety of the user.

\section{conclusion and future work}
We propose LBGP, a follow ahead method that uses both reinforcement learning and point based navigation. We address the limitations of classical methods and end-to-end approaches by combining Deep RL and classical motion planner.
Our implementation outperforms previous work in an obstacle free environment \cite{Nikdel2018TheHP}. To train our deep RL model we used a curriculum learning that gradually increased the difficulty of the person motion model to learn a robust policy for front following. Our results show that using a planner improves the generalizablity and safety of the trained policy compared to an End-to-End method, and allows us to successfully perform zero-shot sim2real transfer.

% Our results show a great improvement in compared to the previous following ahead system
% \cite{Nikdel2018TheHP}.

% benefits from the advantages of classical control and learning-based approaches in a
% way that addresses their individual limitations

In future work, we aim to improve the system by adding obstacles to the environment. We can use other source of information or active user interaction to improve our LBGP system, for instance, anticipate the user's heading from gaze direction.

\bibliographystyle{IEEEtran}
\bibliography{main}

% Generated by IEEEtran.bst, version: 1.14 (2015/08/26)
\begin{thebibliography}{10}
\providecommand{\url}[1]{#1}
\csname url@samestyle\endcsname
\providecommand{\newblock}{\relax}
\providecommand{\bibinfo}[2]{#2}
\providecommand{\BIBentrySTDinterwordspacing}{\spaceskip=0pt\relax}
\providecommand{\BIBentryALTinterwordstretchfactor}{4}
\providecommand{\BIBentryALTinterwordspacing}{\spaceskip=\fontdimen2\font plus
\BIBentryALTinterwordstretchfactor\fontdimen3\font minus
  \fontdimen4\font\relax}
\providecommand{\BIBforeignlanguage}[2]{{%
\expandafter\ifx\csname l@#1\endcsname\relax
\typeout{** WARNING: IEEEtran.bst: No hyphenation pattern has been}%
\typeout{** loaded for the language `#1'. Using the pattern for}%
\typeout{** the default language instead.}%
\else
\language=\csname l@#1\endcsname
\fi
#2}}
\providecommand{\BIBdecl}{\relax}
\BIBdecl

\bibitem{Ho2017}
D.~M. Ho, J.~S. Hu, and J.~J. Wang, ``Behavior control of the mobile robot for
  accompanying in front of a human,'' in \emph{Advanced Intelligent
  Mechatronics (AIM), 2012 IEEE/ASME Int. Conf.}\hskip 1em plus 0.5em minus
  0.4em\relax IEEE, July 2012, pp. 377--382.

\bibitem{Hu2014DesignOS}
J.~Hu, J.-J. Wang, and D.~M. Ho, ``Design of sensing system and anticipative
  behavior for human following of mobile robots,'' \emph{IEEE Transactions on
  Industrial Electronics}, vol.~61, pp. 1916--1927, 2014.

\bibitem{Nikdel2018TheHP}
P.~Nikdel, R.~Shrestha, and R.~Vaughan, ``The hands-free push-cart: Autonomous
  following in front by predicting user trajectory around obstacles,''
  \emph{2018 IEEE International Conference on Robotics and Automation (ICRA)},
  pp. 1--7, 2018.

\bibitem{leigh2015person}
A.~Leigh, J.~Pineau, N.~Olmedo, and H.~Zhang, ``Person tracking and following
  with 2{D} laser scanners,'' in \emph{Robotics and Automation (ICRA), 2015
  IEEE Int. Conf.}\hskip 1em plus 0.5em minus 0.4em\relax IEEE, 2015, pp.
  726--733.

\bibitem{jung2012control}
E.~J. Jung, B.~J. Yi, and S.~Yuta, ``Control algorithms for a mobile robot
  tracking a human in front,'' in \emph{Intelligent Robots and Systems, 2012
  IEEE/RSJ Int. Conf.}\hskip 1em plus 0.5em minus 0.4em\relax IEEE, Oct 2012,
  pp. 2411--2416.

\bibitem{silver2018general}
D.~Silver, T.~Hubert, J.~Schrittwieser, I.~Antonoglou, M.~Lai, A.~Guez,
  M.~Lanctot, L.~Sifre, D.~Kumaran, T.~Graepel \emph{et~al.}, ``A general
  reinforcement learning algorithm that masters chess, shogi, and go through
  self-play,'' \emph{Science}, vol. 362, no. 6419, pp. 1140--1144, 2018.

\bibitem{berner2019dota}
C.~Berner, G.~Brockman, B.~Chan, V.~Cheung, P.~D{\k{e}}biak, C.~Dennison,
  D.~Farhi, Q.~Fischer, S.~Hashme, C.~Hesse \emph{et~al.}, ``Dota 2 with large
  scale deep reinforcement learning,'' \emph{arXiv preprint arXiv:1912.06680},
  2019.

\bibitem{chen2019relational}
C.~Chen, S.~Hu, P.~Nikdel, G.~Mori, and M.~Savva, ``Relational graph learning
  for crowd navigation,'' \emph{arXiv preprint arXiv:1909.13165}, 2019.

\bibitem{kulhanek2019vision}
J.~Kulh{\'a}nek, E.~Derner, T.~de~Bruin, and R.~Babu{\v{s}}ka, ``Vision-based
  navigation using deep reinforcement learning,'' in \emph{2019 European
  Conference on Mobile Robots (ECMR)}.\hskip 1em plus 0.5em minus 0.4em\relax
  IEEE, 2019, pp. 1--8.

\bibitem{Pierre2018EndtoEndDL}
J.~M. Pierre, ``End-to-end deep learning for robotic following,'' in
  \emph{ICMSCE 2018}, 2018.

\bibitem{Wang2018PersonDT}
X.~Wang, L.~Zhang, D.~Wang, and X.~Hu, ``Person detection, tracking and
  following using stereo camera,'' in \emph{International Conference on Graphic
  and Image Processing}, 2018.

\bibitem{Huh2013IntegratedNS}
S.~Huh, D.~Shim, and J.~Kim, ``Integrated navigation system using camera and
  gimbaled laser scanner for indoor and outdoor autonomous flight of uavs,''
  \emph{2013 IEEE/RSJ International Conference on Intelligent Robots and
  Systems}, pp. 3158--3163, 2013.

\bibitem{Lugo2013FrameworkFA}
J.~J. Lugo and A.~Zell, ``Framework for autonomous on-board navigation with the
  ar.drone,'' \emph{Journal of Intelligent \& Robotic Systems}, vol.~73, pp.
  401--412, 2013.

\bibitem{Islam2019TowardAG}
M.~Islam, M.~Fulton, and J.~Sattar, ``Toward a generic diver-following
  algorithm: Balancing robustness and efficiency in deep visual detection,''
  \emph{IEEE Robotics and Automation Letters}, vol.~4, pp. 113--120, 2019.

\bibitem{Zadeh2018OnlinePP}
S.~M. Zadeh, A.~Yazdani, K.~Sammut, and D.~Powers, ``Online path planning for
  auv rendezvous in dynamic cluttered undersea environment using evolutionary
  algorithms,'' \emph{Appl. Soft Comput.}, vol.~70, pp. 929--945, 2018.

\bibitem{Wang2017Realtime3H}
M.~Wang, D.~Su, L.~Shi, Y.~Liu, and J.~V. Mir{\'o}, ``Real-time 3d human
  tracking for mobile robots with multisensors,'' \emph{2017 IEEE International
  Conference on Robotics and Automation (ICRA)}, pp. 5081--5087, 2017.

\bibitem{Goldhoorn2014ContinuousRT}
A.~Goldhoorn, A.~Garrell, R.~Alqu{\'e}zar, and A.~Sanfeliu, ``Continuous real
  time pomcp to find-and-follow people by a humanoid service robot,''
  \emph{2014 IEEE-RAS International Conference on Humanoid Robots}, pp.
  741--747, 2014.

\bibitem{Islam2019PersonfollowingBA}
M.~J. Islam, J.~Hong, and J.~Sattar, ``Person-following by autonomous robots: A
  categorical overview,'' \emph{The International Journal of Robotics
  Research}, vol.~38, pp. 1581 -- 1618, 2019.

\bibitem{Moustris2016IntentionbasedFC}
G.~Moustris and C.~Tzafestas, ``Intention-based front-following control for an
  intelligent robotic rollator in indoor environments,'' \emph{2016 IEEE
  Symposium Series on Computational Intelligence (SSCI)}, pp. 1--7, 2016.

\bibitem{socialaware}
B.~S.~B. {Dewantara} and J.~{Miura}, ``Generation of a socially aware behavior
  of a guide robot using reinforcement learning,'' in \emph{2016 International
  Electronics Symposium (IES)}, 2016, pp. 105--110.

\bibitem{narvekar2019learning}
S.~Narvekar and P.~Stone, ``Learning curriculum policies for reinforcement
  learning,'' in \emph{Proceedings of the 18th International Conference on
  Autonomous Agents and MultiAgent Systems}.\hskip 1em plus 0.5em minus
  0.4em\relax International Foundation for Autonomous Agents and Multiagent
  Systems, 2019, pp. 25--33.

\bibitem{bansal2020combining}
S.~Bansal, V.~Tolani, S.~Gupta, J.~Malik, and C.~Tomlin, ``Combining optimal
  control and learning for visual navigation in novel environments,'' in
  \emph{Conference on Robot Learning}.\hskip 1em plus 0.5em minus 0.4em\relax
  PMLR, 2020, pp. 420--429.

\bibitem{Li2019}
A.~Li, S.~Bansal, G.~Giovanis, V.~Tolani, C.~Tomlin, and M.~Chen, ``{Generating
  Robust Supervision for Learning-Based Visual Navigation Using Hamilton-Jacobi
  Reachability},'' in \emph{Conference on Learning for Dynamics and Control},
  2019.

\bibitem{d4pg}
G.~Barth-Maron, M.~W. Hoffman, D.~Budden, W.~Dabney, D.~Horgan, A.~Muldal,
  N.~Heess, and T.~Lillicrap, ``Distributed distributional deterministic policy
  gradients,'' \emph{arXiv preprint arXiv:1804.08617}, 2018.

\bibitem{redmon2016yolo9000}
J.~Redmon and A.~Farhadi, ``{YOLO9000: Better, Faster, Stronger},'' \emph{arXiv
  preprint arXiv:1612.08242}, 2016.

\bibitem{quigley2009ros}
M.~Quigley, K.~Conley, B.~Gerkey, J.~Faust, T.~Foote, J.~Leibs, R.~Wheeler, and
  A.~Y. Ng, ``{ROS}: an open-source {R}obot {O}perating {S}ystem,'' in
  \emph{ICRA workshop on open source software}, vol.~3, no. 3.2.\hskip 1em plus
  0.5em minus 0.4em\relax Kobe, 2009, p.~5.

\bibitem{1389727}
N.~{Koenig} and A.~{Howard}, ``Design and use paradigms for gazebo, an
  open-source multi-robot simulator,'' in \emph{2004 IEEE/RSJ International
  Conference on Intelligent Robots and Systems (IROS) (IEEE Cat.
  No.04CH37566)}, vol.~3, 2004, pp. 2149--2154 vol.3.

\end{thebibliography}

\end{document}